\documentclass[a4paper]{article}
\usepackage[margin=25mm]{geometry}
\usepackage{amsmath}
\usepackage{amsfonts}
\usepackage{amssymb}
\usepackage{graphicx}
\usepackage{verbatim}
\immediate\write18{texcount -tex -sum  \jobname.tex > \jobname.wordcount.tex}

\providecommand{\keywords}[1]
{
	\small	
	\textbf{\textit{Keywords---}} #1
}

\title{On Error Correction Neural Networks for Economic Forecasting\footnote{© 2020 IEEE.  Personal use of this material is permitted.  Permission from IEEE must be obtained for all other uses, in any current or future media, including reprinting/republishing this material for advertising or promotional purposes, creating new collective works, for resale or redistribution to servers or lists, or reuse of any copyrighted component of this work in other works.}}
\author{Mhlasakululeka Mvubu$^{1,2}$, Emmanuel Kabuga$^{1,3}$, Christian Plitz$^{4}$, \\
	Bubacarr Bah$^{1,2}$,  Ronnie Becker$^{1}$, Hans Georg Zimmermann$^{5}$ \\
	\small $^{1}$AIMS South Africa \\
	\small $^{2}$Stellenbosch University \\
	\small $^{3}$University of Cape Town \\
	\small $^{4}$Technical University of Munich \\
	\small $^{5}$Fraunhofer Society \\
}
\date{} 

\begin{document}
\maketitle

\begin{abstract}
	Recurrent neural networks (RNNs) are more suitable for learning non-linear dependencies in dynamical systems from observed time series data. In practice, all the external variables driving such systems are not known a priori, especially in economical forecasting. A class of RNNs called Error Correction Neural Networks (ECNNs) was designed to compensate for missing input variables. It does this by feeding back in the current step the error made in the previous step.  
	The ECNN is implemented in Python by the computation of the appropriate gradients and it is tested on stock market predictions. As expected it outperformed the simple RNN and LSTM and other hybrid models which involve a de-noising pre-processing step. The intuition for the latter is that de-noising may lead to loss of information. 
\end{abstract} \hspace{10pt}


\section{Introduction}
\label{sec:intro}

\subsection{Recurrent Neural Networks}
\label{sec:rnn}
Most applications like economics and finance have non-linear dependencies that a better modelled by RNNs with their universal approximation properties \cite{b27}. Time series modelling and natural language modelling are two good examples where RNNs are dominantly used. Unlike feedforward neural networks, RNNs have feedback connections in which outputs of the model are fed back into itself. RNNs proposed in \cite{b1} are specifically designed to handle sequential data and to exploit dependences in such type of data. In other words RNNs model dynamical systems, with typically the following state relations.\vskip -2mm
\begin{equation}
\label{eqn:dsys}
s^{(t)} = f\left(s^{(t-1)}, x^{(t-1)};\theta\right) ,
\end{equation}\vskip -2mm
\noindent where $s^{(t)}$ is the state of the system at time $t$, which is a function $f$ of the previous state $s^{(t-1)}$ and an external signal $x^{(t-1)}$ at time $t-1$, parametrized by $\theta$. 

The simple (standard or vanilla) RNN has $T$ inputs (data points) $\left\{x^{(t)}\right\}_{t=1}^{T}$ with corresponding outputs $\left\{y^{(t)}\right\}_{t=2}^{T+1}$. Note that each $x^{(t)}$ and $y^{(t)}$ maybe a vector. Key in the success of RNNs, that is their ability to generalized well, is due to the fact that it is trained using parameter sharing. We denote parameter (weight) matrices that connect states (hidden layers) as $A$, those that connect inputs to states as $B$, and those that connect states to outputs as $C$. The dynamical system that the simple RNN represents is described by the following non-linear system of equations.\vskip -2mm
\begin{align}
\label{eqn:rnn-se}
\mbox{state equation:} \qquad ~s^{(t)} &=	f\left(As^{(t-1)} + Bx^{(t-1)}\right),\vspace{-2mm}\\
\label{eqn:rnn-oe}
\mbox{output equation:} \qquad ~y^{(t)} &= g\left(Cs^{(t)}\right),
\end{align}\vskip -1mm
\noindent where $f$ and $g$ are activation functions. 

RNNs are trained using back-propagation through time (BPTT). BPTT is a natural extension of standard back-propagation performing gradient descent on a network unfolded in time. More details on BPTT will be given in Section \ref{sec:ecnn-imp}. Having complete knowledge of all external drivers of a dynamical system is the underlying assumption of most RNNs. In which there is a reliance on the architecture of the RNN with its long-term memory to learn the dynamics of the system.

\subsection{Error Correction Neural Networks}
\label{sec:ecnn}
\begin{figure}[htbp]
	\vskip -2mm
	\begin{center}
		\includegraphics[width=0.75\columnwidth,height=0.3\columnwidth]{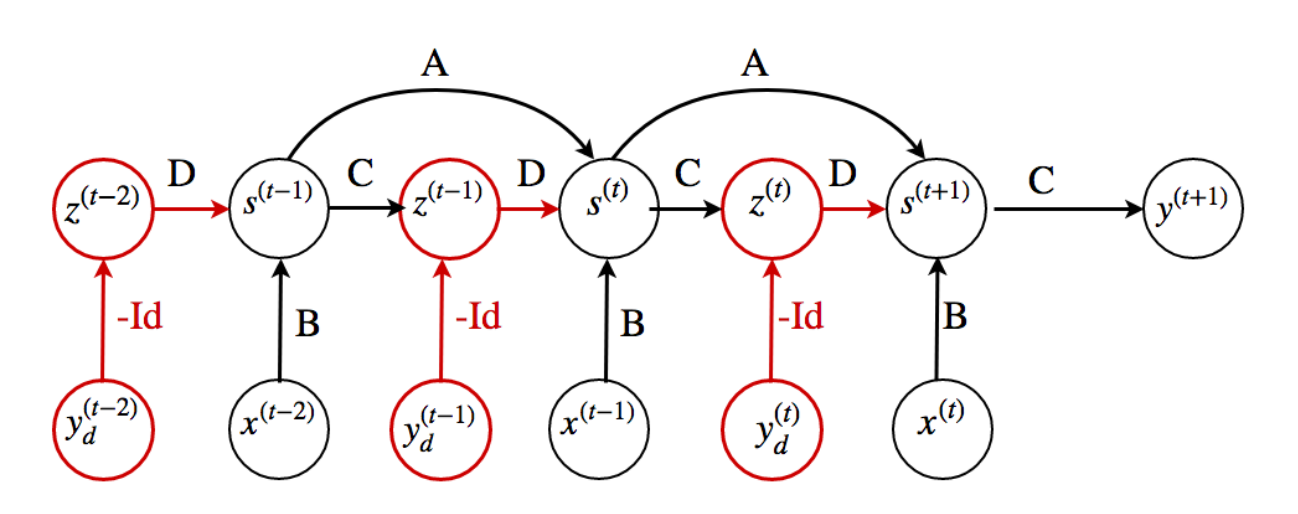}
		\caption{Unfolded ECNN where the weight matrices $A, B, C,$ and $D$ are shared weights and $-Id$ is the negative identity matrix. An ECNN whose only recurrence is the feedback connection from the error $z^{(t-2)}$ calculated from the previous time step $t-2$ to output to the hidden layer. At each time step $t$, the input is $x^{(t)}$, the states are $s^{(t)}$, the outputs are $y^{(t+1)}$.}
		\label{fig:ecnn-cgraph}
	\end{center}
	\vskip -0.5cm
\end{figure}

The ECNN was designed in \cite{b5} with the awareness that not all external variables of the dynamical system are known in practice. This is its main difference with the standard RNN, LSTM, and GRU. The ECNN feeds back into itself the prediction error made in the previous time. The error essentially tells us something about the missing external variables. This is similar to Auto Regressive Integrated Moving Average (ARIMA) models except that ARIMA models are linear, thus making them an inadequate framework for non--linear dynamical systems \cite{b28}. Denoting the target output at time $t$ as ${y}_d^{(t)}$ and the error in the previous time step as $y^{(t-1)} - {y}_d^{(t-1)}$, the governing equations of the ECNN as follows.
\begin{align}
\label{eqn:ecnn-ee}
z^{(t)} &=	y^{(t)} - {y}_d^{(t)},\vspace{-2mm}\\
\label{eqn:ecnn-se}
s^{(t)} &=	\tanh\left(As^{(t-1)} + Bx^{(t-1)} + D\tanh\left(z^{(t-1)}\right)\right),\vspace{-2mm}\\
\label{eqn:ecnn-oe}
y^{(t)} &= Cs^{(t)}.
\end{align}\vskip -2mm
\noindent Note that the error, $z^{(t)}$ goes through a non-linearity in \eqref{eqn:ecnn-se}. Otherwise, the $D$ could be absorbed into the $A$ and $B$.
The computational graph of the ECNN is given Figure \ref{fig:ecnn-cgraph}.

\subsection{Time Series Forecasting using Deep Learning}
\label{sec:stock}

There is quite a bit of work in predicting time series using neural networks. In general, all kinds of neural networks have been used in the application of deep learning to financial time series. Artificial neural networks were used in \cite{b7}; convolutional neural networks were used in \cite{b23}; deep belief networks in \cite{b24,b25}; recurrent neural networks in \cite{b5}\cite{b8}; and stacked autoencoders in \cite{b6}. This work builds on the work of \cite{b5}\cite{b8} for the ECNN and that of  \cite{b6} for the comparison of the ECNN with existing RNN based methods for stock market prediction. 

Precisely, in \cite{b6}  a new deep learning framework was presented where wavelet transforms (WT), stacked autoencoders (SAEs), and LSTM are combined for stock price forecasting. Their framework develops a six years predictive performance of the four proposed models in different stock markets, WT-LSTM (combining WT and LSTM), WSAEs-LSTM (combining SAEs and WT-LSTM), RNN, and LSTM. The WSAEs-LSTM model has SAEs-LSTM as the main part of their model and is used to learn the deep features of financial time series, while WT is used as a preprocessing step that attempts to de-noise the input. We benchmarked this work with the results of \cite{b6}.

\subsection{Contributions}
\label{sec:contr}
The contribution of this work is three-fold. Firstly, it implemented the ECNN by deriving the gradients that make the BPTT possible, see Section \ref{sec:ecnn-imp}. As hinted above the ECNN was introduced in \cite{b5} but the implementation has been proprietary and hence the deep learning and the financial forecasting communities are largely not aware of the existence of the ECNN. Following the publication of this paper an open source package in Python would be made available.

Secondly, to test the ECNN, a comparison of its performance was made against existing RNN models and hybrid models proposed in \cite{b6} on 4 popular stock indices: S\&P 500 (GSPC) Index, Hang Seng Index (HSI), Nikkei 225 Index (N225), and Dow Jones Industrial Average (DJIA) Index. Results show the superior performance of the ECNN over the other models. 

Thirdly, on the question of smoothing and non-smoothing, the ECNN shows good ability to learn from real data, which is typically noisy -- suggesting that smoothing might not necessarily be the best idea.


\section{ECNN Implementation}
\label{sec:ecnn-imp}
The ECNN learning takes place by finding the optimal parameters that minimize the empirical loss. Formally, learning implies solving the following optimization problem.
\begin{equation}
\min_{A, B, C, D} ~ L := \frac{1}{T}\sum_{t=1}^{T} L^{(t)},
\end{equation}
where $L^{(t)} = \ell \left(y^{(t)}, {y}_d^{(t)}\right)$ is the loss at time $t$. 

\subsection{Back-Propagation Through-Time}
\label{sec:bptt}
ECNN learning takes place by finding the optimal parameters that minimize the empirical loss. Formally, learning implies solving the following optimization problem.
\begin{equation}
\min_{A, B, C, D} ~ L := \frac{1}{T}\sum_{t=1}^{T} L^{(t)},
\end{equation}
where $L^{(t)} = \ell \left(y^{(t)}, {y}_d^{(t)}\right)$ is the loss at time $t$. 

\subsection{Back-Propagation Through-Time}
\label{sec:bptt}
The ECNN is implemented using BPTT, which uses gradient descent or its variants. Individually, each parameter (denoted here as a dummy $W$) is updated as follows.
\begin{equation}
W^{(j+1)} = W^{(j)} - \eta \nabla_W L|_{W=W^{(j)}}, \quad j = 0, 1, 2, \ldots
\end{equation} 
where $\eta$ is referred to as a the {\em learning rate} and $j$ is the iteration counter.

BPTT iterates over the following steps until convergence.
\begin{enumerate}
	\item {\bf Forward pass}: step through $k$ time steps, compute the hidden and output states.\vspace{1mm}
	
	\item Compute the loss, summed over the previous time steps.\vspace{1mm}
	
	\item Compute the gradient of the loss function w.r.t. to all the parameters over the previous $k$ time steps. \vspace{1mm}
	
	\item {\bf Backward pass}: update parameters.
\end{enumerate}

\subsection{Computing Gradients}
\label{sec:gradients}
Computing gradients is a key step in the BPTT algorithm above, hence computing gradients was a key component of the implementation of the ECNN. As a result, this task is characterized as one of the main contributions of this work. For easy readability, below we state the gradients w.r.t. each weight matrix and the detailed derivations are shown in Section \ref{sec:deriv} of the Appendix.
\begin{align}
\label{eqn:gradA}
\nabla_A L =& \sum_{t=1}^{T} \mbox{diag}\left({\bf 1}-\tilde{s}^{(t)}\right)\left(\nabla_{s^{(t)}}L \right) s^{(t-1)^T} \\
\label{eqn:gradB}
\nabla_B L =& \sum_{t=1}^{T} \mbox{diag}\left({\bf 1}-\tilde{s}^{(t)}\right)\left(\nabla_{s^{(t)}}L \right) x^{(t-1)^T}  \\
\label{eqn:gradC}
\nabla_C L =& \sum_{t=1}^{T} \mbox{diag}\left({\bf 1}-\tilde{z}^{(t)}\right) \cdot D^T \mbox{diag}\left({\bf 1}-\tilde{s}^{(t)}\right) C^T \left(\nabla_{y^{(t)}}L \right) s^{(t-1)^T} + ~\sum_{t=1}^{T} \left(\nabla_{y^{(t)}}L \right) s^{(t)^T} \\
\label{eqn:gradD}
\nabla_D L =& \sum_{t=1}^{T}\mbox{diag}\left({\bf 1}-\tilde{s}^{(t)}\right)\left(\nabla_{s^{(t)}}L \right) z^{(t-1)^T}
\end{align}
where ${\bf 1}$ is a vector of ones; $\tilde{q}$ is a vector with the components the vector $q$ squared, i.e $\tilde{q}_i = q_i^2$; $v^T$ denotes the transpose of $v$; and $\mbox{diag}(u)$ denotes a diagonal matrix with the vector $u$ on the diagonal. $\nabla_{s^{(t)}}L$ and $\nabla_{y^{(t)}}L$ are derived in Section \ref{sec:deriv}.

\section{Evaluating Performance of ECNN}
\label{sec:stocks}

\subsection{Experiment Setting}
\label{sec:exp}
\begin{figure}[htbp]
	\vskip -2mm
	\begin{center}
		\includegraphics[width=0.75\columnwidth,height=0.35\columnwidth]{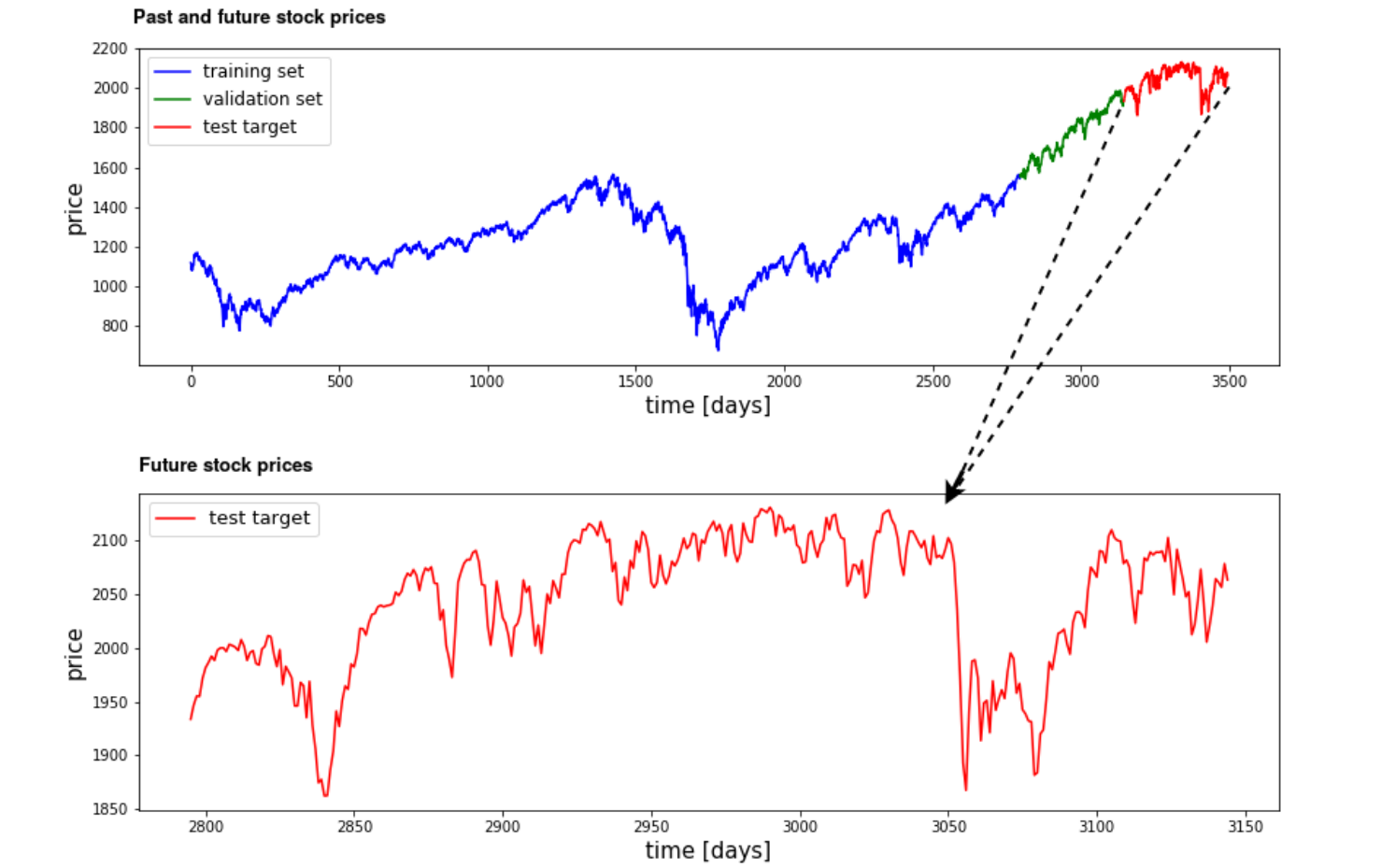}
		\caption{S\&P 500 daily closing price development between the period of 2002-01-01 and 2015-12-30. Plots also showing the split of the data into training, validation, and testing.}
		\label{fig:data-split}
	\end{center}
	\vskip -0.5cm
\end{figure}

\paragraph{Data}
The ECNN algorithm was tested in the forecasting of daily movement (the algorithm also works for predicting more than one day) for four stock indices in different stock markets: S\&P 500, HSI, N225, and DJIA. For each market, ECNN was compared to standard RNN (denoted RNN), and LSTM. Models were trained over historical stock data from Yahoo Finance \cite{b20} for the period from 2002-01-01 to 2016-01-01. A glimpse into these datasets is given in Figure \ref{fig:data-split}. Each database entry of the stock index stated above includes closing price, opening price, high price (signifying highest price of the day), low price (signifying lowest price of the day), and volume of trade. In addition to these historical stock trading data, technical indicators of a stock trade are also used to improve the prediction, \cite{b21}. The most popular indicators are {\em Moving Average} (MA), {\em Exponential Moving Average} (EMA), {\em Moving Average Convergence Divergence} (MACD), {\em Average True Range} (ATR) and {\em Stochastic} ($\% K$). The mathematical expressions for these indicators are in Section \ref{sec:tech} of the Appendix.

\paragraph{Training and Testing}
To improve the predictive performance of the model during training \cite{b22}, the data is normalized as follows:
\begin{equation}
\hat{x}^{(t)} = \frac{x^{(t)} - \min\left(x^{(t)}\right)}{\max\left(x^{(t)}\right) - \min\left(x^{(t)}\right)}
\end{equation}
Similarly from $y_d^{(t)}$ to $\hat{y}_d^{(t)}$. 

In our experiments, we took different rolling window sizes. For example, for a rolling window of 10 days of historical prices: $\hat{x}^{(i)} = \left\{ \hat{x}^{(t-9,i)}, \hat{x}^{(t-8,i)}, \ldots, \hat{x}^{(t,i)} \right\}$, and corresponding output is $y_d^{(t+1,i)} \in \mathbb{R}$. The mean squared error (MSE) at time $t+1$ is used for the training loss, $L^{(t+1)}$.
\begin{equation}
L^{(t+1)} = \frac{1}{2K}\sum_{t=T-N+1}^{T}\sum_{i\in U_t}\left(\hat{y}_d^{(t,i)} - \hat{y}^{(t,i)}\right)^2,
\end{equation}
where $K$ is the number of all training examples, $N$ is the rolling window size, $U_t$ is the input data universe at $t$, and $\hat{y}^{(t,i)} = f\left(\hat{x}^{(t,i)};W^{(t+1)}\right)$ with $W^{(t+1)}$ as the parameters.

The data was split into training (80\%), validation (10\%) and testing (10\%). An example can be seen from Figure \ref{fig:data-split}. Most recent data were selected for testing, next recent dataset was selected for the validation and rest were selected for the training. We used three classical performance criteria \cite{b15,b15,b17,b18} to measure the predictive accuracy of each model. These are the {\em Relative measure} (Theil U), {\em Mean Absolute Percentage Error} (MAPE) and {\em Pearson’s Correlation Coefficient} (R). For details see Section \ref{sec:perform} of the Appendix. Most training was done on the Centre for High Performance Computing (CHPC) \cite{b26}.

Another performance criteria is the {\em directional accuracy} (DA), which measures the proportion of the days when the model forecasts the correct direction of price movement and is defined as thus:
\begin{equation}
\frac{1}{m-1} \sum_{t=1}^{m-1} pos{\left(\left(y_d^{(t+1)} - y_d^{(t)}\right)\left(y^{(t+1)} - y^{(t)}\right)\right)}
\end{equation}
where $pos$ is an unary operator defined as:
\begin{equation}
pos(x)=
\begin{cases}
1 ~~\mbox{if} ~x>0\\
0 ~~\mbox{otherwise}.
\end{cases}
\end{equation}

\paragraph{Trading Strategy}
As part of the experiments, a trading strategy was investigated. The direction of a price development is relevant to traders, as the direction may directly determine whether to take a long or a short position in the market. A trading strategy is a way to buy and sell in markets based on predefined rules for trading decisions. Trading strategies are employed to avoid behavioural finance biases and ensure consistent results. We consider the strategy of buying and selling based on the predicted results of each model. The strategy recommends that investors buy when the next period’s expected value is higher than the actual value; while it recommends that investors sell when the predicted value at day $t$ is smaller than the predicted value at day $t+1$. More precisely, the strategy can be described by the following equations:
\begin{align}
y^{(t+1)} &> y^{(t)} \quad : \quad \mbox{Buy} \\
y^{(t+1)} &< y^{(t)} \quad : \quad \mbox{Sell}
\end{align}
where $y^{(t)}$ is the current predicted closing price and $y^{(t+1)}$ is the predicted closing price for the following market day.

The total return of this trading rule can be calculated using the following equation and used as a scale of comparison between models and markets:
\begin{multline}
R = 100\cdot\left(\sum_{t=1}^{b}\frac{y^{(t+1)} - y^{(t)} + \left(S\cdot y^{(t+1)} + B\cdot y^{(t)}\right)}{y^{(t)}} + \sum_{t=1}^{s}\frac{y^{(t+1)} - y^{(t)} + \left(B\cdot y^{(t+1)} + S\cdot y^{(t)}\right)}{y^{(t)}}\right)
\end{multline}
where $R$ is the strategy returns, $b$ and $s$ denote the total number of days for buying and selling, respectively \cite{b6,b19}, $B$ and $S$ are the transaction costs for buying and selling, respectively. Here we choose the unified cost in the spot as 0.25\% for buying and 0.45\% for selling as used in \cite{b6}.

\subsection{Results}
\label{sec:results}
\begin{figure}[h!]
	\vskip -0mm
	\centering
	\includegraphics[width=0.53\columnwidth]{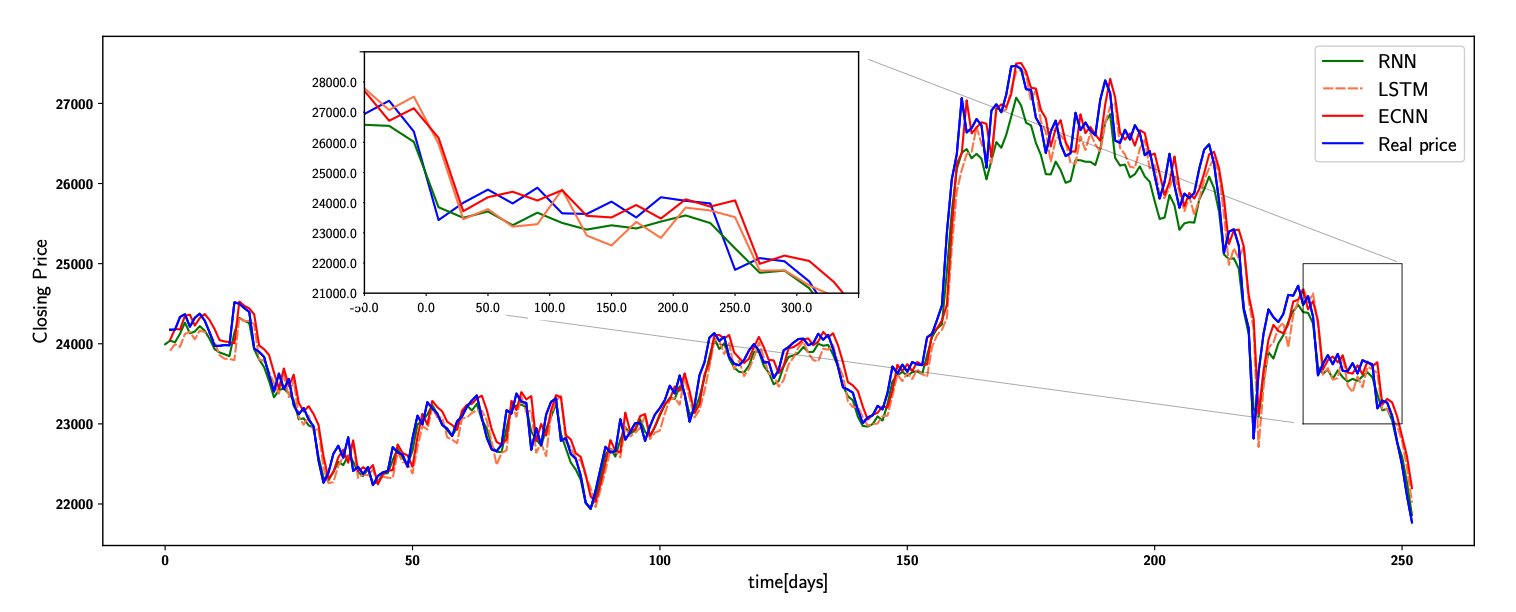}
	\includegraphics[width=0.45\columnwidth]{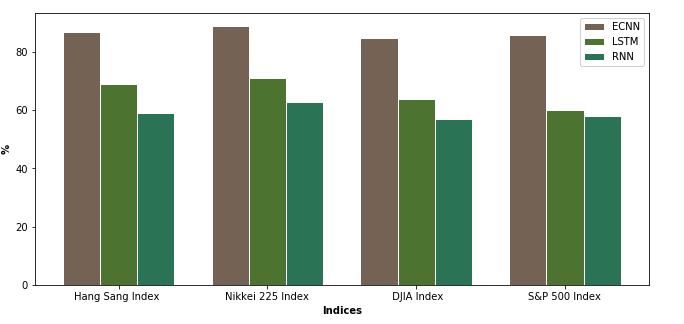}\vspace{-2mm}
	\caption{({\em Left:})Comparisons of actual data and predicted HSI data using RNN, LSTM and ECNN with a
		rolling window size of 60. ({\em Right:})Average yearly directional accuracies (in \%) of the ECNN, LSTM and RNN for the predicted indices.}
	\label{fig:hsi-price}
\end{figure}

The performance of ECNN was better than both LSTM and RNN. The left panel of Figure \ref{fig:hsi-price} shows plots of sample results for the HSI test data. More plots in Section \ref{sec:extra} of the Appendix.  Note that each algorithm is trained with its optimal hyper-parameters, through a search in hyper-parameter space.

We investigated the predictive accuracy of the 3 algorithms using the 4 performance criteria outlined in Section \ref{sec:exp}. We further compared our results to the results of two other algorithms proposed in \cite{b6}. These are hybrid algorithms: (i) WLSTM, and (ii) WSAEs-LSTM. For a fair comparison, we used 4 out of the 6 stocks used on in \cite{b6}.  
We report average annual accuracies, as well as average 6--year accuracies using the 6 years of test data.
\begin{table}[h!]
	\caption{Return of the models for the predicted Nikkei 225 Index.}
	\vskip 1mm
	\centering
	\begin{tabular}{|c|cccccc|c|}
		\hline
		{\bf Model} & {\bf Year 1} & {\bf Year 2} & {\bf Year 3} & {\bf Year 4} & {\bf Year 5} & {\bf Year 6} & {\bf Average}\\
		\hline
		\hline
		{\bf RNN} & -5.84 & -3.98 & -14.53 & 9.52 & 52.26 & 25.35 & 10.47\\\
		{\bf LSTM} & 18.08 & 33.35 & -14.84 & 55.43 & 7.73 & 24.93 & 20.78\\
		{\bf ECNN} & {\bf 70.99} & {\bf 41.96} & {\bf 60.33} & {\bf 82.76} & {\bf 49.93} & {\bf 73.98} & {\bf 63.33}\\
		{\bf buy-\&-hold} & -18.49 & 3.16 & 53.29 & 11.94 & 5.44 & -8.58 & 7.79\\
		\hline
	\end{tabular}
	\vspace{1mm}
	\label{tab:n225-profit}
	\vskip -0cm
\end{table}

\vspace{-0mm}\begin{table}[h!]
	\caption{Predictive accuracy for DJIA of the different models using performance criteria: ({\em top}) MAPE, ({\em middle}) R, ({\em bottom}) Theil U.}
	\vskip 2mm
	\centering
	\begin{tabular}{|c|cccccc|c|} 
		\hline
		{\bf Model} & {\bf Year 1} & {\bf Year 2} & {\bf Year 3} & {\bf Year 4} & {\bf Year 5} & {\bf Year 6} & {\bf Average}\\
		\hline
		\hline
		{\bf RNN} & 0.034 & 0.038 & 0.031 & 0.028 & 0.031 & 0.037 & 0.033\\
		{\bf LSTM} & 0.020 & 0.028 & 0.019 & 0.023 & 0.017 & 0.023 & 0.022\\
		{\bf ECNN} & {\bf 0.008} & {\bf 0.010} &  0.012 & 0.009 & 0.011 & {\bf 0.010} & {\bf 0.010}\\
		{\bf WLSTM \cite{b6}} & 0.015 & 0.018 & 0.013 & 0.011 & 0.017 & 0.012 & 0.014\\
		{\bf WSAEs-LSTM \cite{b6}} & 0.016 & 0.013 & {\bf 0.009} & {\bf 0.008} & {\bf 0.008} & {\bf 0.010} & 0.011\\
		\hline
	\end{tabular}
	\vspace{2mm}\\
	\begin{tabular}{|c|cccccc|c|} 
		\hline
		{\bf Model} & {\bf Year 1} & {\bf Year 2} & {\bf Year 3} & {\bf Year 4} & {\bf Year 5} & {\bf Year 6} & {\bf Average}\\
		\hline
		\hline
		{\bf RNN} & 0.681 & 0.591 & 0.814 & 0.681 & 0.513 & 0.419 & 0.607\\
		{\bf LSTM} & 0.831 & 0.761 & 0.733 & 0.891 & 0.912 & 0.737 & 0.811\\
		{\bf ECNN} & {\bf 0.932} & {\bf 0.941} & 0.930 & {\bf 0.981} & {\bf 0.973} & {\bf 0.954} & {\bf 0.952}\\
		{\bf WLSTM \cite{b6}} & 0.915 & 0.871 & 0.963 & 0.911 & 0.817 & 0.927 & 0.901\\
		{\bf WSAEs-LSTM \cite{b6}} & 0.922 & 0.928 & {\bf 0.984} & 0.952 & 0.953 & 0.952 & 0.949\\
		\hline
	\end{tabular}
	\vspace{2mm}\\
	\begin{tabular}{|c|cccccc|c|} 
		\hline
		{\bf Model} & {\bf Year 1} & {\bf Year 2} & {\bf Year 3} & {\bf Year 4} & {\bf Year 5} & {\bf Year 6} & {\bf Average}\\
		\hline
		\hline
		{\bf RNN} & 0.031 & 0.026 & 0.018 & 0.015 & 0.023 & 0.018 & 0.022\\
		{\bf LSTM} & 0.012 & 0.015 & 0.013 & 0.013 & 0.016 & 0.014 & 0.014\\
		{\bf ECNN} & {\bf 0.007} & {\bf 0.005} & 0.010 & {\bf 0.004} & 0.006 & 0.007 & {\bf 0.007}\\
		{\bf WLSTM \cite{b6}} & 0.010 & 0.012 & 0.008 & 0.007 & 0.011 & 0.008 & 0.009\\
		{\bf WSAEs-LSTM \cite{b6}} & 0.010 & 0.009 & {\bf 0.006} &  0.005 & {\bf 0.005} & {\bf 0.006} & {\bf 0.007}\\
		\hline
	\end{tabular}
	\vspace{-0.2cm}
	\label{tab:djia}
\end{table}

\vspace{0.0mm}
The ECNN is doing much better than RNN, LSTM, and WLSTM but performing as good, if not better than, as WSLSTM, without the preprocessing in WLSTM and WSLSTM. See results in Table \ref{tab:djia} for the DJIA index in the 3 performance criteria; while similar results for the other 3 stocks are in Section \ref{sec:extra} of the Appendix. 

The results of the predictive DA for each model are shown in the right panel of Figure \ref{fig:hsi-price}. It can be seen that ECNN out performs RNN and LSTM.

Finally, for the outcome of the trading strategy, we report annual returns of the six years of test data for the 3 algorithms, RNN, LSTM, and ECNN. These results are also compared to the returns of a buy-and-hold strategy. Tables \ref{tab:n225-profit} and \ref{tab:gspc-profit} confirms the better performance of ECNN over LSTM, RNN and buy-and-hold on the N225 and S\&P 500 indices respectively. Plots for the other 2 stocks are in Section \ref{sec:extra} of the Appendix.

\begin{table}[h!]
	\caption{Return of the models for the predicted S\&P 500 Index.}
	\vskip 1mm
	\centering
	\begin{tabular}{|c|cccccc|c|}
		\hline
		{\bf Model} & {\bf Year 1} & {\bf Year 2} & {\bf Year 3} & {\bf Year 4} & {\bf Year 5} & {\bf Year 6} & {\bf Average}\\
		\hline
		\hline
		{\bf RNN} & 17.44 & 12.35 & 15.83 & -9.53 & -11.36 & 6.37 & 5.18\\
		{\bf LSTM} & -5.93 & 19.37 & 26.98 & -1.94 & 3.83 & 44.27 & 14.43\\
		{\bf ECNN} & {\bf 79.36} & {\bf 61.81} & {\bf 49.26} & {\bf 43.26} & {\bf 83.94} & {\bf 81.03} & {\bf 49.08}\\
		{\bf buy-\&-hold} & -8.38 & -6.38 & 20.37 & 10.31 & -6.83 & 10.19 & 6.65\\
		\hline
	\end{tabular}
	\vspace{1mm}
	\label{tab:gspc-profit}
	\vskip -0.2cm
\end{table}


\section{Extensions}
\label{sec:extensions}

As a form of extension, we investigate the impact of adding a smoothing pre-processing step to the ECNN. We used one of the most popular smoothing techniques, that exponential smoothing.

\subsection{ECNNs with Exponential Smoothing}
\label{sec:expo}

Exponential smoothing is a technique developed in the late 1950s and early 1960s {\cite{b14}}, {\cite{b9}} and {\cite{b13}}. It enhances the accuracy of a correct prediction for short-term forecasting like it is shown in {\cite{b10}}. The original exponential smoothing algorithms work on univariate data but progress has been made to use exponential smoothing also with multivariate data {\cite{b11}}. In practice, the technique is smoothing data from time series and it thus works as a window function. Unlike other window functions, the exponential smoothing (ES) - algorithm is using exponentially decreasing weights for the time series data. Depending on the parameter, an emphasis is put on data multiple time steps in the past or on more recent time steps. It is shown in Smyl's state-of-the-art and M4 competition-winning approach {\cite{b12}} that the ES algorithm is useful to capture the main components like seasonality and level of a time series. Here is the smoothing formula for non-seasonal data.

\begin{equation}\label{eq1}
l_t = \alpha y_t + (1 - \alpha)l_{t-1}
\end{equation}

The level $l_t$ is calculated for every time step of the series, while $y_t$ is the value of the series at the specified time step. The parameter $\alpha\in (0,1)$ is the smoothing coefficient. Although it could technically also assigned to $\alpha = 0$, there is no value in it since the level $l_t$ would be a constant. If we set $\alpha = 1$ on the other hand, $l_t$ would just be the value of the series again which is not smoothing our data either. 

After calculating the seasonality and the level of the data, we want to apply the obtained time series to our original data to get the input $X_t$:
\begin{align}
X_t = \ln(Input_t/l_t) \label{eq5}
\end{align}
This time series $X_t$ is normalized and squashed due to the $\ln$ function so it can be used as an input for the ECNN. The output of the ECNN is  
\begin{align}
O_t = ECNN(X_t). \label{eq6}
\end{align}
Rescaling of $X_t$ leads to the predicted time series data 
\begin{align}
\hat{Y_t} = l_t\cdot \exp({O_t}). \label{eq7}
\end{align}

\subsection{Experimental Results}
\label{sec:experiment}

The datasets were checked for missing values and normalized as shown in the previous section. The Hang Seng Index dataset was used to compare LSTM and ECNN including exponential smoothing and the ECNN without exponential smoothing in regards to computational time and accuracy.

For the hyper-parameter selection of the exponential smoothing formula, we chose $\alpha=0.8$ as it got us good results. For the LSTM and ECNN, we used different hyper-parameter for batch size, time step (rolling window), neurons (units used in the ECNN layer), and the learning rate. The optimal hyper-parameters can be seen in Table \ref{tab:hyperp}.

\begin{table}[htbp]
	\caption{Optimal hyper-parameters used in the experiment.}
	\vskip 2mm
	\centering
	\begin{tabular}{|c|c|c|c|}
		\hline
		{\bf Epochs} & {\bf Time step} & {\bf Neurons} & {\bf Learning rate} \\
		\hline
		1000 & 7 & 32 & $10^{-3}$ \\
		\hline
		\hline
		{\bf Batch size}: & ECNN = 64 & LSTM ES = 8 & ECNN ES = 32 \\
		\hline
	\end{tabular}
	\label{tab:hyperp}
\end{table}

The selection was performed using experimentation. The batch size values and the number of neurons tried were among the following $\left\{2^k\right\}_{k=0}^{10}$. The time step values were between 5 and 20. The learning rate values tried were in the range of $10^{-2}$ and $10^{-6}$. All the models were run over 1000 epochs and the model parameters were updated using the Adam optimizer.

\begin{table}[h]
	\caption{Comparison of the 3 models: ECNN, ECNN with exponential smoothing, and LSTM with exponential smoothing}
	\centering
	\begin{tabular}{|l|r|r|r|}
		\hline
		Model 	& ECNN 	& LSTM ES & ECNN ES \\
		\hline
		$R^2$  	& 0.989612 & 0.999619 & 0.999621\\ 
		\hline
	\end{tabular}
	\label{tab:performance}
\end{table}
In terms of the $R^2$ score the performance of the ECNN essentially remains the same when exponential smoothing is used, see Table \ref{tab:performance}. The explanation is that the ECNN can squeeze out the noise where it does not contain useful information and keeps the `noise' that contain useful information. hence blanket smoothing or de-noising may lead to loss of information.

Moreover, the downside of smoothing is that the computational effort is increased especially for LSTM, see Table \ref{tab:runtime}, where the run time is measured in seconds. The experiment is run on a local machine and on Google Colaboratory (Colab) The local machine uses an 2.6GHz Intel Core i7-6700HQ (quad-core, up to 3.5GHz with Turbo Boost) as CPU, 16GB of RAM and an Nvidia GeForce GTX 1060 (6GB GDDR5 VRAM) graphic card. Google Colab runs with a Nvidia Tesla T4 (16 GB GDDR6 VRAM) GPU and an Intel(R) Xeon(R) CPU @ 2.00GHz CPU.
\begin{table}[h]
	\caption{Runtime comparison of the three models, locally and on Google Colab. Time in seconds}
	\vskip 0.1cm
	\centering
	\begin{tabular}{|l|r|r|}
		\hline
		Model & Runtime local & Runtime cloud \\
		\hline
		ECNN        & 2162.91 & 271.49 \\
		LSTM ES     & 16364.86 & 3521.72 \\
		ECNN ES     & 1704.60 & 440.76\\
		\hline
	\end{tabular}
	\label{tab:runtime}
\end{table}

\section{Conclusion}
\label{sec:concl}

This research implemented the ECNN in Python and offered evidence for the favourable performance on the predictive accuracy and profitability of returns of ECNN over other models on Hong Kong’s Hang Seng index, Japan’s NIKKEI 225 index, and the USA's S\&P 500 and DJIA indices from 2010 to 2016. These experiments demonstrates the ECNN's ability to model unknown external forces of a dynamical system by the incorporation of the error it is making as an input the system, where the error serves as a proxy for the missing external input variables of the system. The performance is mainly attributed to the nature of financial markets, which typify forecasting where only a part of the external variables is known or observable. Furthermore, we show that universal approximators like the ECNN can learn high dimensional non-linear relationships from real world noisy data without the need for smoothing. Smoothing will destroy information about micro-structures in economics time series data.


\section{Appendix}
\label{sec:appdx}

\subsection{Detailed Derivations}
\label{sec:deriv}

We need to compute the gradient of our loss function with respect to the weight matrices
$A$, $B$, $C$, $D$. The parameter $C$ is present in two functions which are \textit{state transition} with the \textit{external force} and \textit{output equation} as shown by \eqref{eqn:ecnn-se}. Lets consider the network parameters in Figure \ref{fig:ecnn-cgraph} as the set $\{ \beta = A, B, C, D \}$, and $s^{(t)}$ as the hidden state of the network at time $t$, we can write the gradients as

\begin{equation} 
\label{eq8}
\frac{\partial L}{\partial \beta} = \sum_{t=1}^{T} \frac{\partial L^{(t)}}{\partial \beta}
\end{equation}
where $L^{(t)}$ is the loss at time $t$. The expansion of loss function gradients at time $t$ is
\begin{equation} 
\label{eq9}
\frac{\partial L^{(t)}}{\partial \beta} = \sum_{k=1}^{t} \frac{\partial L^{(t)}}{\partial y^{(t)}} \frac{\partial y^{(t)}}{\partial s^{(t)}}\frac{\partial s^{(t)}}{\partial s^{(k)}}\frac{\partial s^{(k)}}{\partial \beta}
\end{equation}
This shows how parameters in the set $\beta$ affect the loss function at the previous time-steps (i.e. $k < t$).
 
The gradient $\nabla_{y^{(t)}} L$ on the outputs at time step $t$, for all $i$, $t$, is as follows:
\begin{equation}
\label{eq10}
\left(\nabla_{y^{(t)}} L\right)_{i} = \frac{\partial L^{t}}{\partial y_{i}^{(t)}} = y_{i}^{(t)} - \hat{y}_i^{(t)}.
\end{equation}
Going backward, starting from the end of the sequence, at the final time step $T$, $s^{(T)}$ only has $y^{(T)}$ as a descendent, so its gradient is simple:
\begin{equation}
\nabla_{s^{(T)}} L = \left(\nabla_{s^{(T)}} y^{(T)}\right)^{T} \nabla_{y^{(T)}} L = C^{T} \nabla_{y^{(T)}} L
\label{eq11}
\end{equation}
where $\nabla_{y^{(T)}} L = y^{(T)}-\hat{y}^{(T)}$, $\hat{y}^{(T)}$ and $y^{(T)}$ are the target and predicted values at the final time step $T$ respectively. For each time $t < T$, in decreasing order of $t$, we proceed backward along each path starting at $s^{(t+1)}$ and finishing at $s^{(t)}$ and also starting at the $L^{(T)}$ proceed backward similarly until reaching $s^{(t)}$. This gives $\frac{\partial L}{\partial s^{(t)}}$ as follows:
\begin{equation}
\small \nabla_{s^{(t)}} L = \left(\frac{\partial s^{(t+1)}}{\partial s^{(t)}}\right)^{T}(\nabla_{s^{(t+1)}} L) + \left(\frac{\partial y^{(t)}}{\partial s^{(t)}}\right)^{T}\left(\nabla_{y^{(t)}} L\right). \label{eq12}
\end{equation}

Now at time step $t + 1$ (\ref{eqn:ecnn-se}) can be written as
\begin{equation}
\small s^{(t+1)} =	\tanh\left(As^{(t)} + Bx^{(t)} + D\tanh(C s^{(t)}-y_d^{(t)})\right). \label{eq13}
\end{equation}

This enables us to obtain our Jacobian matrix for the hidden state parameter as
\begin{align}
\left(\frac{\partial s^{(t+1)}}{\partial s^{(t)}}\right)^{T} 
=& 
\left(A^{T} + \left(\frac{\partial}{\partial s^{(t)}} \tanh\left(C s^{(t)} - \hat{y}^{(t)}\right)\right)^{T} D^{T}\right) \times \mbox{diag}\left({\bf 1}-\tilde{s}^{(t+1)}\right),  \nonumber \\
=& \left(A^{T} + \left(\mbox{diag}\left(1-\tilde{z}^{(t)}\right) C\right)^{T} D^{T}\right) \times \mbox{diag}\left({\bf 1}-\tilde{s}^{(t+1)}\right) \nonumber \\
=& \left(A^{T} + C^{T} \mbox{diag}\left(1-\tilde{z}^{(t)}\right) D^{T}\right) \times \mbox{diag}\left({\bf 1}-\tilde{s}^{(t+1)}\right) \label{eq14}
\end{align}
where (here and the sequel) $\tilde{u}$ is a vector with the components of vector $u$ squared, i.e $\tilde{u}_i = u_i^2$.
Also $\left(\frac{\partial y^{(t)}}{\partial s^{(t)}}\right)^{T}$ is given by
\begin{equation}
\left(\frac{\partial y^{(t)}}{\partial s^{(t)}}\right)^{T}= C^{T} \label{eq15}
\end{equation}

Substituting (\ref{eq14}) and (\ref{eq15}) into (\ref{eq12}) gives
\begin{align}
\nabla_{s^{(t)}} L =&  \left[\left(A^{T} + C^{T} \mbox{diag}\left({\bf 1}-\tilde{z}^{(t)}\right) D^{T}\right) \times \mbox{diag}\left({\bf 1}-\tilde{s}^{(t+1)}\right) + C^{T}\right]\left(\nabla_{s^{(t+1)}} L\right) \label{eq16}
\end{align}

For parameter $A$, it appears in the arguments for both $s^{(t)}$ and $y^{(t)}$, so we will have to check in both $s^{(t)}$ and $y^{(t)}$. We also make note that $y^{(t)}$ depends on $A$ both directly and indirectly (through $s^{(t-1)}$). Using notation described in Section \ref{sec:bptt}, the gradient on the remaining parameters is given by:
\begin{equation}
\nabla_{A} L =  \sum_{t}\sum_{i}\left(\frac{\partial L}{\partial s_{i}^{(t)}}\cdot \frac{\partial s_{i}^{(t)}}{\partial A}\right) \label{eq17}
\end{equation}

The final term, however, requires us to notice that there is an implicit dependence of $s^{(t)}$ on $A_{ij}$ through $s^{(t-1)}$ as well as a direct dependence. Now, the derivative of the internal hidden state w.r.t. weight matrix $A$ is
\begin{equation}
\frac{\partial s^{(t)}}{\partial A} = \mbox{diag}\left({\bf 1}-\tilde{s}^{(t)}\right)s^{(t-1)^{T}}.\label{eq18}
\end{equation}
Then we substitute (\ref{eq18}) into  (\ref{eq17}) to get \eqref{eqn:gradA} where $\nabla_{s^{(t)}} L$ is described by (\ref{eq12}). Taking the gradient of $B$ and $D$ is similar to doing it for $A$ since they both require taking sequential derivatives of $s^{(t)}$. Hence for $B$ and $D$ we get \eqref{eqn:gradB} and \eqref{eqn:gradD}.
To find the derivative of the loss function w.r.t. parameter $C$, we firstly define $\nabla_C L$ as
\begin{align}
\nabla_{C} L =  \sum_{t}\sum_{i}(\nabla_{y_{i}^{(t)}}L) (\nabla_{C}y_{i}^{(t)})\label{eq22}
\end{align}

Since $y^{(t)} = Cs^{(t)}$, then (\ref{eq22}) becomes
\begin{align}
\nabla_{C} L =&  \sum_{t}\sum_{i}\left(\nabla_{y_{i}^{(t)}}L\right) \nabla_{C}(Cs^{(t)})_{i} \nonumber \\ 
=& \sum_{t}(\nabla_{y^{(t)}}L)s^{(t)^{T}} + \sum_{t}\sum_{i}\sum_{j}C_{ij}(\nabla_{y_{i}^{(t)}}L)(\nabla_{C}s^{(t)})_{j}. \nonumber
\label{eq23}
\end{align}
which simplifies to \eqref{eqn:gradC}.

\subsection{Technical Indicators}
\label{sec:tech}
The descriptions of the technical indicators are as follows:
\paragraph{Moving Average}
A moving average ($MA$) is a trend indicator that dynamically calculates the mean average of prices over a defined number of past, defined as follows.
$$MA = CCP - OCP$$
where $CCP$ and $OCP$ are current closing and old closing price for a predetermined period (5 and 10 days).

\paragraph{Exponential Moving Average}
Exponential Moving Average ($EMA$) for a predetermined period (20 days) in each point is calculated according to the following formula:
$$	EMA^{(t)} = EMA^{(t-1)} + \alpha\left(x^{(t)} - EMA^{(t-1)}\right)$$
where $\alpha = \frac{2}{n+1}$, $n$ is the length of the $EMA$, $x^{(t)}$ is the current closing price, $EMA^{(t-1)}$ is the previous $EMA$ value, and $EMA^{(t)}$ is the current $EMA$ value.

\paragraph{Moving Average Convergence Divergence}
It, for a predetermined period (12 and 26 days), is constructed based on exponential moving averages. It is calculated by subtracting the longer exponential moving average ($EMA$) of window length $N$ from the shorter $EMA$ of window length $M$, where the $EMA$ is computed as follows:
$$ EMA^{(t)}(N) = EMA^{(t-1)}(N) + \frac{2}{N}\left(P^{(t)} - EMA^{(t-1)(N)}\right) $$
where $EMA^{(t)}(N)$ is the exponential moving average at time $t$, $N$ is the window length of the $EMA$, and $P^{(t)}$ is the value of index at time $t$.

\paragraph{Average True Range}
The average true range ($ATR$) is a measure of price volatility. $ATR$ is calculated by selecting 14 days, which is commonly used and by adding the true range of the present day to that of the previous 13 days, then dividing by 14. This would be your initial $ATR$:
$$ATR = ( 13 \mbox{ previous } ATR + \mbox{today’s true range } ) /14.$$

\paragraph{Stochastic Fast (\%K)}
The stochastic fast ( \%K ) is a momentum indicator comparing the closing price of a security to the range of its prices over a certain period and is defined by the following expression:
$$\%K = (CCP - L14) / (H14 - L14) $$
where $L14$ and $H14$ denote the lowest low and highest high of the past 14 days respectively.

\subsection{Performance Criteria}
\label{sec:perform}
The definitions of performance criteria are presented below.

\paragraph{Theil U}
It is the relative measure of the difference between two variables. It squares the deviation to give more weight to large errors and exaggerates errors. Theil U is defined as follows:
$$ \mbox{Theil U} = \frac{\sqrt{\frac{1}{N}\sum_{t=1}^{N}\left(\hat{y}^{(t)} - y^{(t)}\right)^2}}{\sqrt{\frac{1}{N}\sum_{t=1}^{N}\left(\hat{y}^{(t)}\right)^2} + \sqrt{\frac{1}{N}\sum_{t=1}^{N}\left(y^{(t)}\right)^2}},$$
where $N$ is the number of predictions and $\hat{y}^{(t)}$ and $y^{(t)}$ are predicted and actual values, respectively.

\paragraph{Mean Absolute Percentage Error}
The {\em Mean Absolute Percentage Error} (MAPE) is used as a predictive accuracy measure to determine which model performs the best \cite{b22}. MAPE can evaluate and compare models' predictive power and is defined as follows.
$$\mbox{MAPE} =\frac{1}{N}\sum_{t=1}^{N}\left|\left(\hat{y}^{(t)} - y^{(t)} \right) / y^{(t)}\right|,$$
where $\hat{y}^{(t)}$ is the target value and $y^{(t)}$ is the predicted value. A lower MAPE value indicates better network performance, but we cannot expect it to be close to zero, as financial markets are so volatile and fluctuating.  

\paragraph{Pearson’s correlation coefficient}
{\em Pearson’s correlation coefficient} (R) is the measurement of the linear correlation between two variables. Large R values mean more shared variation which means more accurate predictions are possible about one variable based on nothing more than knowledge of the other variable. R is defined as follows:
$$\mbox{R} = \frac{\sum_{t=1}^{N}\left({y}^{(t)} - \overline{y^{(t)}}\right)\left(\hat{y}^{(t)} - \overline{\hat{y}^{(t)}}\right)} {\sqrt{\sum_{t=1}^{N}\left({y}^{(t)} - \overline{y^{(t)}}\right)^2\left(\hat{y}^{(t)} - \overline{\hat{y}^{(t)}}\right)^2}}$$
where $\hat{y}^{(t)}$ and $y^{(t)}$ are predicted and actual values respectively and $N$ represents the prediction period.


\subsection{Extra Tables}
\label{sec:extra}
\begin{figure}[h!]
	\centering
	\includegraphics[width=0.48\columnwidth]{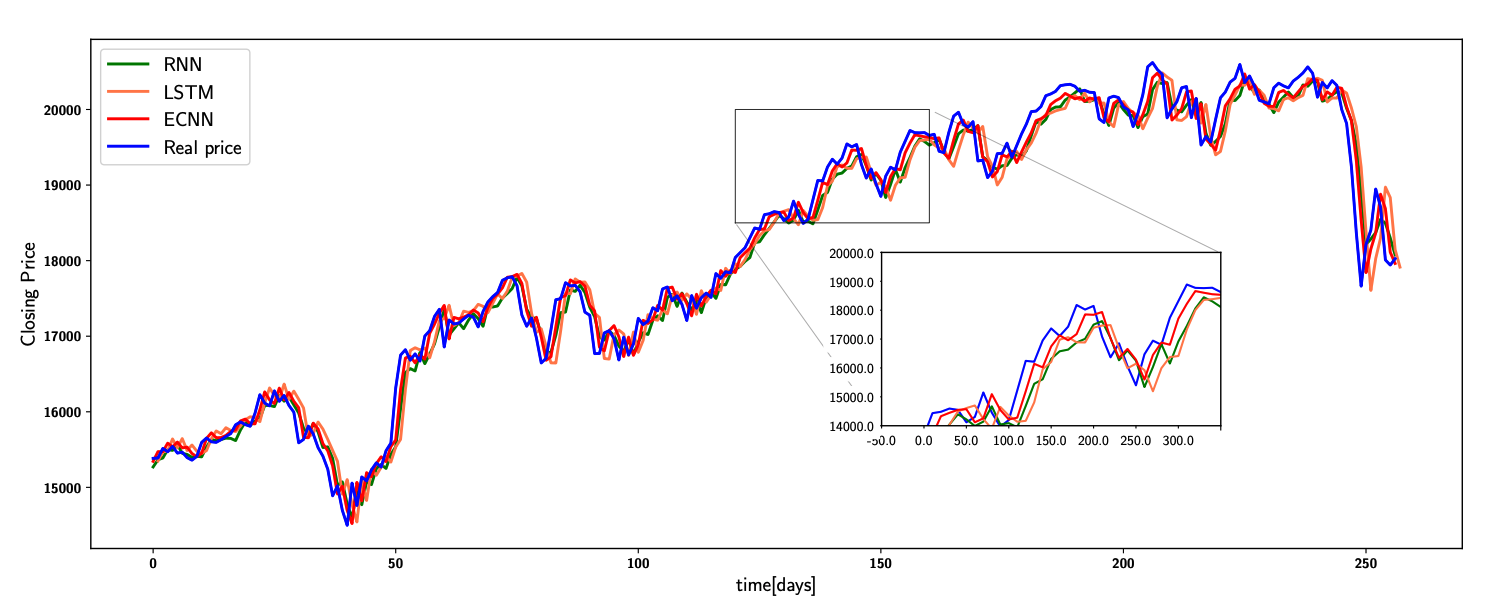}
	\includegraphics[width=0.48\columnwidth]{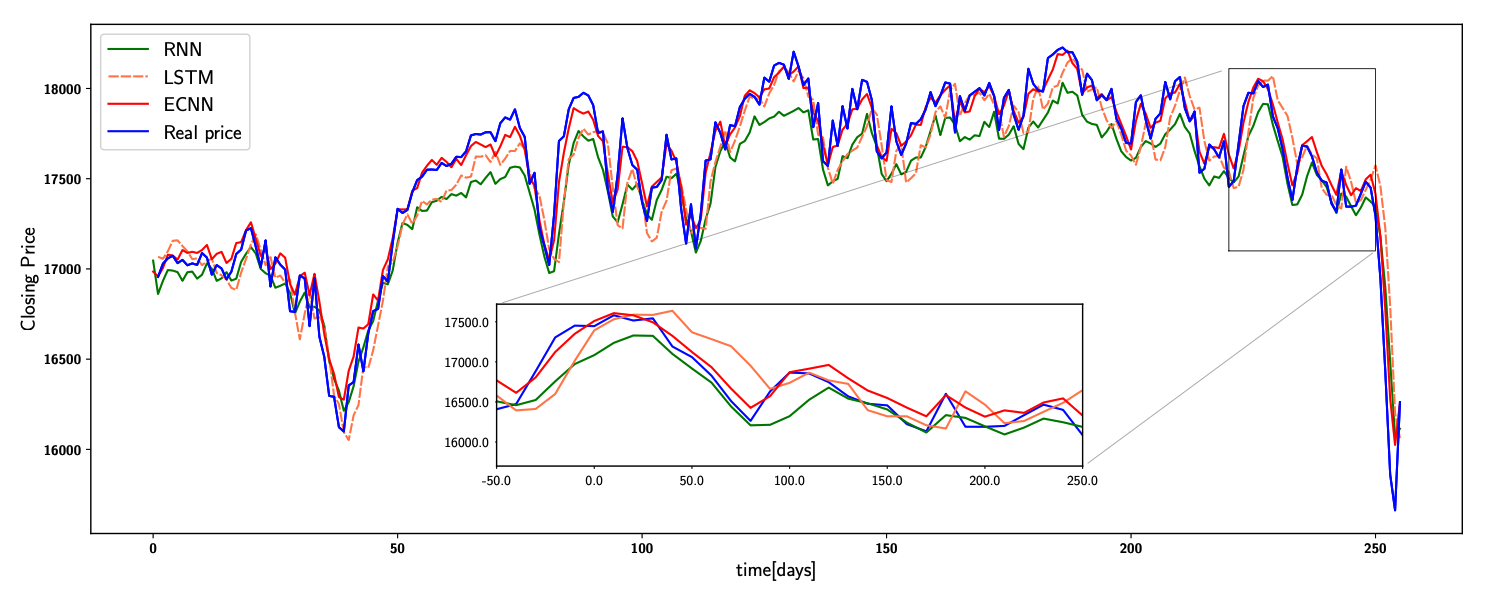}
	\includegraphics[width=0.48\columnwidth]{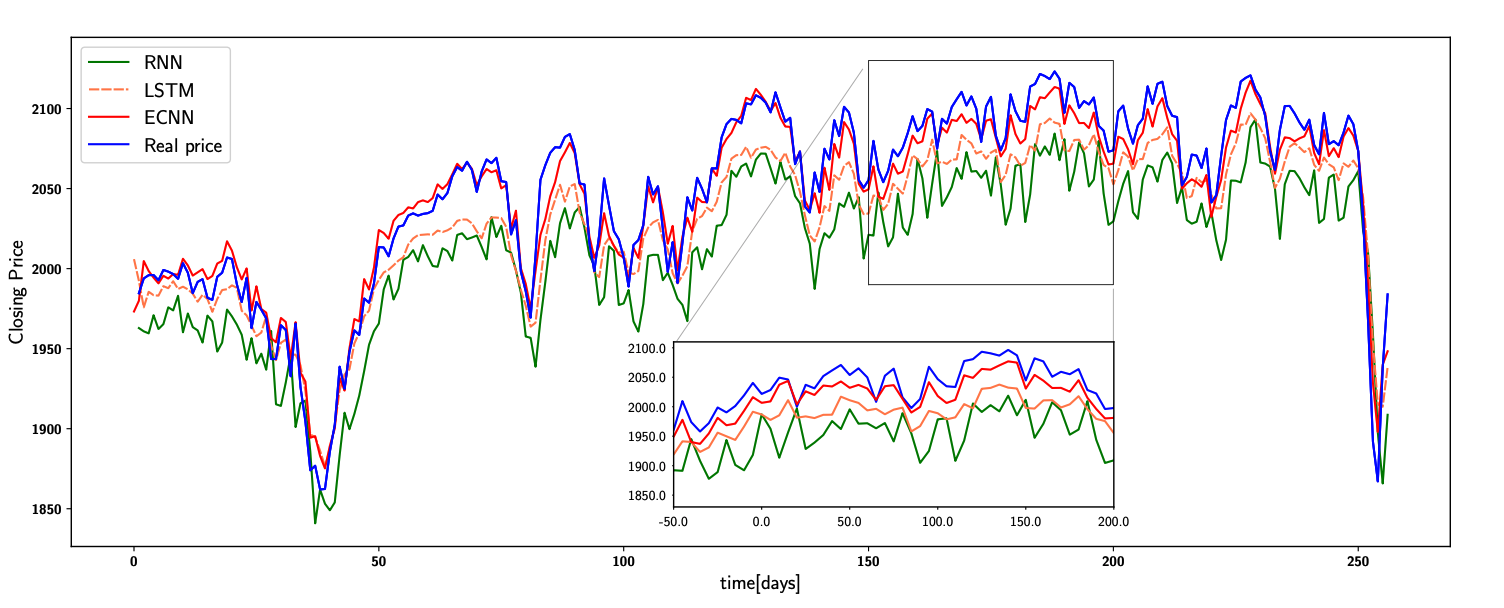}\vspace{-2mm}
	\caption{Comparisons of actual and predictions using RNN, LSTM and ECNN with a
		rolling window of 60, ({\em top left}) N225, ({\em top right}) DJIA, ({\em bottom}) S\&P 500.}
	\label{fig:n225+djia+gspc}
	\vspace{-1mm}
\end{figure}

More plots of sample results of the rolling window of 60 for the test data of N225, DJIA, and S\&P 500 shown in Figure \ref{fig:n225+djia+gspc}.

More tables depicting how the ECNN outperforms the RNN, and LSTM, and compares favourably with WLSTM and WSLSTM. Tables \ref{tab:hsi}, \ref{tab:n225}, and \ref{tab:gspc} for the HSI, N225, and S\&P 500 indices respectively in the 3 performance criteria (i.e. MAPE, R, and Theil U).

\vspace{0mm}\begin{table}[h!]
	\caption{Predictive accuracy for HSI of the different models using performance criteria: ({\em top}) MAPE, ({\em middle}) R, ({\em bottom}) Theil U.}
	\vspace{1mm}
	\centering
	\begin{tabular}{|c|cccccc|c|} 
		\hline
		{\bf Model} & {\bf Year 1} & {\bf Year 2} & {\bf Year 3} & {\bf Year 4} & {\bf Year 5} & {\bf Year 6} & {\bf Average}\\
		\hline
		\hline
		{\bf RNN} & 0.031 & 0.038 & 0.034 & 0.040 & 0.033 & 0.030 & 0.034\\
		{\bf LSTM} & 0.027 & 0.030 & 0.026 & 0.021 & 0.024 & 0.023 & 0.025\\
		{\bf ECNN} & {\bf 0.013} & {\bf 0.015} & {\bf 0.015} & {\bf 0.011} & {\bf 0.012} & {\bf 0.011} & {\bf 0.013}\\
		{\bf WLSTM \cite{b6}} & 0.020 & 0.027 & 0.017 & 0.018 & 0.028 & 0.021 & 0.022\\
		{\bf WSAEs-LSTM \cite{b6}} & 0.016 & 0.017 & 0.017 & {\bf 0.011} & 0.021 & 0.013 & 0.015\\
		\hline
	\end{tabular}
	\vspace{3mm}\\
	\begin{tabular}{|c|cccccc|c|} 
		\hline
		{\bf Model} & {\bf Year 1} & {\bf Year 2} & {\bf Year 3} & {\bf Year 4} & {\bf Year 5} & {\bf Year 6} & {\bf Average}\\
		\hline
		\hline
		{\bf RNN} & 0.782 & 0.631 & 0.671 & 0.812 & 0.677 & 0.791 & 0.727\\
		{\bf LSTM} & 0.863 & 0.866 & 0.815 & 0.910 & 0.795 & 0.841 & 0.848\\
		{\bf ECNN} & {\bf 0.951} & {\bf 0.966} & {\bf 0.931} & {\bf 0.953} & {\bf 0.967} & {\bf 0.981} & {\bf 0.958}\\
		{\bf WLSTM \cite{b6}} & 0.935 & 0.810 & 0.858 & 0.833 & 0.900 & 0.917 & 0.876\\
		{\bf WSAEs-LSTM \cite{b6}} & 0.944 & 0.924 & 0.920 & {0.927} & 0.904 & 0.968 & 0.931\\
		\hline
	\end{tabular}
	\vspace{3mm}\\
	\begin{tabular}{|c|cccccc|c|}
		\hline
		{\bf Model} & {\bf Year 1} & {\bf Year 2} & {\bf Year 3} & {\bf Year 4} & {\bf Year 5} & {\bf Year 6} & {\bf Average}\\
		\hline
		\hline
		{\bf RNN} & 0.023 & 0.024 & 0.019 & 0.022 & 0.021 & 0.019 & 0.021\\
		{\bf LSTM} & 0.014 & 0.011 & 0.013 & 0.016 & 0.015 & 0.015 & 0.014\\
		{\bf ECNN} & {\bf 0.009} & 0.011 & {\bf 0.006} & {\bf 0.006} & {\bf 0.007} & {\bf 0.008} & {\bf 0.008}\\
		{\bf WLSTM \cite{b6}} & 0.012 & 0.017 & 0.011 & 0.011 & 0.021 & 0.013 & 0.014\\
		{\bf WSAEs-LSTM \cite{b6}} & 0.011 & {\bf 0.010} & 0.008 & 0.007 & 0.018 & {\bf 0.008} & 0.011\\
		\hline
	\end{tabular}
	\vspace{-0.5mm}
	\label{tab:hsi}
\end{table}

\begin{table}[h!]
	\caption{Predictive accuracy for N225 of the different models using performance criteria: ({\em top}) MAPE, ({\em middle}) R, ({\em bottom}) Theil U.}
	\vskip 1mm
	\centering
	\begin{tabular}{|c|cccccc|c|} 
		\hline
		{\bf Model} & {\bf Year 1} & {\bf Year 2} & {\bf Year 3} & {\bf Year 4} & {\bf Year 5} & {\bf Year 6} & {\bf Average}\\
		\hline
		\hline
		{\bf RNN} & 0.043 & 0.037 & 0.033 & 0.035 & 0.038 & 0.029 & 0.036\\
		{\bf LSTM} & 0.037 & 0.030 & 0.029 & 0.031 & 0.030 & 0.023 & 0.030\\
		{\bf ECNN} & {\bf 0.012} & {\bf 0.011} & {\bf 0.016} & {\bf 0.011} & {\bf 0.015} & {\bf 0.017} & {\bf 0.014}\\
		{\bf WLSTM \cite{b6}} & 0.033 & 0.025 & 0.032 & 0.025 & 0.022 & 0.027 & 0.028\\
		{\bf WSAEs-LSTM \cite{b6}} & 0.020 & 0.016 & 0.017 & 0.014 & 0.016 & 0.018 & 0.017\\
		\hline
	\end{tabular}
	\vspace{3mm}\\
	\begin{tabular}{|c|cccccc|c|} 
		\hline
		{\bf Model} & {\bf Year 1} & {\bf Year 2} & {\bf Year 3} & {\bf Year 4} & {\bf Year 5} & {\bf Year 6} & {\bf Average}\\
		\hline
		\hline
		{\bf RNN} & 0.631 & 0.694 & 0.893 & 0.779 & 0.832 & 0.833 & 0.780\\
		{\bf LSTM} & 0.773 & 0.877 & 0.927 & 0.883 & 0.910 & 0.818 & 0.865\\
		{\bf ECNN} & {\bf 0.899} & {\bf 0.933} & 0.989 & {\bf 0.991} & {\bf 0.986} & {\bf 0.963} & {\bf 0.960}\\
		{\bf WLSTM \cite{b6}} & 0.748 & 0.838 & 0.973 & 0.786 & 0.951 & 0.906 & 0.867\\
		{\bf WSAEs-LSTM \cite{b6}} & 0.895 & 0.927 & {\bf 0.992} & 0.885 & 0.974 & 0.951 & 0.937\\
		\hline
	\end{tabular}
	\vspace{3mm}\\
	\begin{tabular}{|c|cccccc|c|} 
		\hline
		{\bf Model} & {\bf Year 1} & {\bf Year 2} & {\bf Year 3} & {\bf Year 4} & {\bf Year 5} & {\bf Year 6} & {\bf Average}\\
		\hline
		\hline
		{\bf RNN} & 0.028 & 0.023 & 0.021 & 0.020 & 0.025 & 0.024 & 0.022\\
		{\bf LSTM} & 0.019 & 0.018 & 0.021 & 0.023 & 0.019 & 0.020 & 0.032\\
		{\bf ECNN} & {\bf 0.011} & {\bf 0.009} & {\bf 0.010} & {\bf 0.008} & {\bf 0.009} & {\bf 0.011} & {\bf 0.010}\\
		{\bf WLSTM \cite{b6}} & 0.021 & 0.017 & 0.021 & 0.015 & 0.013 & 0.017 & 0.018\\
		{\bf WSAEs-LSTM \cite{b6}} & 0.013 & 0.010 & {\bf 0.010} & 0.009 & 0.010 & {\bf 0.011} & 0.011\\
		\hline
	\end{tabular}
	\label{tab:n225}
	\vskip -0.5cm
\end{table}

\begin{table}[h!]
	\caption{Predictive accuracy for S\&p 500 (GSPC) of the different models using performance criteria: ({\em top}) MAPE, ({\em middle}) R, ({\em bottom}) Theil U.}
	\vskip 0mm
	\centering
	\begin{tabular}{|c|cccccc|c|} 
		\hline
		{\bf Model} & {\bf Year 1} & {\bf Year 2} & {\bf Year 3} & {\bf Year 4} & {\bf Year 5} & {\bf Year 6} & {\bf Average}\\
		\hline
		\hline
		{\bf RNN} & 0.024 & 0.015 & 0.031 & 0.015 & 0.031 & 0.036 & 0.025\\
		{\bf LSTM} & 0.031 & 0.023 & 0.011 & 0.029 & 0.033 & 0.021 & 0.024\\
		{\bf ECNN} & {\bf 0.009} & {\bf 0.013} & 0.013 & {\bf 0.007} & 0.015 & 0.012 & {\bf 0.011}\\
		{\bf WLSTM \cite{b6}} & 0.015 & 0.020 & 0.012 & 0.010 & 0.015 & 0.015 & 0.015\\
		{\bf WSAEs-LSTM \cite{b6}} & 0.012 & 0.014 & {\bf 0.010} & 0.008 & {\bf 0.011} & {\bf 0.010} & {\bf 0.011}\\
		\hline
	\end{tabular}
	\vspace{3mm}\\
	\begin{tabular}{|c|cccccc|c|} 
		\hline
		{\bf Model} & {\bf Year 1} & {\bf Year 2} & {\bf Year 3} & {\bf Year 4} & {\bf Year 5} & {\bf Year 6} & {\bf Average}\\
		\hline
		\hline
		{\bf RNN} & 0.874 & 0.915 & 0.673 & 0.779 & 0.821 & 0.787 & 0.801\\
		{\bf LSTM} & 0.883 & 0.921 & 0.891 & 0.941 & 0.960 & 0.811 & 0.901\\
		{\bf ECNN} & {\bf 0.966} & {\bf 0.985} & 0.972 & 0.891 & {\bf 0.974} & {\bf 0.981} & {\bf 0.962}\\
		{\bf WLSTM \cite{b6}} & 0.917 & 0.886 & 0.971 & 0.957 & 0.772 & 0.860 & 0.894\\
		{\bf WSAEs-LSTM \cite{b6}} & 0.944 & 0.944 & {\bf 0.984} & {\bf 0.973} & 0.880 & 0.953 & 0.946\\
		\hline
	\end{tabular}
	\vspace{3mm}\\
	\begin{tabular}{|c|cccccc|c|} 
		\hline
		{\bf Model} & {\bf Year 1} & {\bf Year 2} & {\bf Year 3} & {\bf Year 4} & {\bf Year 5} & {\bf Year 6} & {\bf Average}\\
		\hline
		\hline
		{\bf RNN} & 0.012 & 0.013 & 0.012 & 0.008 & 0.011 & 0.014 & 0.012\\
		{\bf LSTM} & 0.007 & 0.011 & 0.009 & 0.010 & {\bf 0.008} & 0.010 & 0.014\\
		{\bf ECNN} & {\bf 0.006} & {\bf 0.005} & 0.007 & 0.008 & {\bf 0.008} & {\bf 0.006} & {\bf 0.007}\\
		{\bf WLSTM \cite{b6}} & 0.011 & 0.014 & 0.008 & 0.007 & 0.011 & 0.011 & 0.010\\
		{\bf WSAEs-LSTM \cite{b6}} & 0.009 & 0.010 & {\bf 0.006} & {\bf 0.005} & {\bf 0.008} & {\bf 0.006} & {\bf 0.007}\\
		\hline
	\end{tabular}
	\vspace{1mm}
	\label{tab:gspc}
\end{table}

\pagebreak
With regards to the profitability test of the trading strategy, Tables \ref{tab:hsi-profit} and \ref{tab:djia-profit} confirming the better performance of ECNN over LSTM, RNN and buy-and-hold on the HSI and DJIA indices respectively. 
\vspace{-2mm}\begin{table}[h!]
	\caption{Return of the models for the predicted HSI.}
	\vskip 2mm
	\centering
	\begin{tabular}{|c|cccccc|c|}
		\hline
		{\bf Model} & {\bf Year 1} & {\bf Year 2} & {\bf Year 3} & {\bf Year 4} & {\bf Year 5} & {\bf Year 6} & {\bf Average}\\
		\hline
		\hline
		{\bf RNN} & 7.35 & 44.03 & -14.33 & -17.33 & 38.34 & 41.05 & 16.52\\
		{\bf LSTM} & 25.57 & 22.76 & 3.85 & 15.14 & {\bf 63.30} & 27.44 & 26.34\\
		{\bf ECNN} & {\bf 80.86} & {\bf 62.74} & {\bf 77.45} & {\bf 57.54} & 53.59 & {\bf 69.23} & {\bf 66.90}\\
		{\bf buy-\&-hold} & -29.62 & 19.34 & 11.31 & 9.65 & -1.78 & -2.02 & 1.15\\
		\hline
	\end{tabular}
	\label{tab:hsi-profit}
	\vskip -0.1cm
\end{table}

\vspace{-2mm}\begin{table}[h!]
	\caption{Return of the models for the predicted DJIA Index.}
	\vskip 2mm
	\centering
	\begin{tabular}{|c|cccccc|c|}
		\hline
		{\bf Model} & {\bf Year 1} & {\bf Year 2} & {\bf Year 3} & {\bf Year 4} & {\bf Year 5} & {\bf Year 6} & {\bf Average}\\
		\hline
		\hline
		{\bf RNN} & 8.36 & 3.44 & 11.12 & 6.36 & -3.45 & 40.26 & 11.02\\
		{\bf LSTM} & 44.28 & 31.44 & 21.38 & -6.35 & 7.38 & 41.37 & 23.25\\
		{\bf ECNN} & {\bf 79.36} & {\bf 61.81} & {\bf 49.26} & {\bf 43.26} & {\bf 83.94} & {\bf 81.03} & {\bf 66.45}\\
		{\bf buy-\&-hold} & -8.38 & -6.38 & 20.37 & 10.31 & -6.83 & 10.19 & 3.22\\
		\hline
	\end{tabular}
	\vspace{1mm}
	\label{tab:djia-profit}
\end{table}



\end{document}